\documentclass{article}

\usepackage{PRIMEarxiv}

\usepackage[utf8]{inputenc} 
\usepackage[T1]{fontenc}    
\usepackage{hyperref}       
\usepackage{url}            
\usepackage{booktabs}       
\usepackage{amsmath}
\usepackage{amssymb}
\usepackage{booktabs}
\usepackage{multirow}
\usepackage{amsfonts}       
\usepackage{nicefrac}       
\usepackage{microtype}      
\usepackage{lipsum}
\usepackage{fancyhdr}       
\usepackage{graphicx}       
\usepackage{caption}
\usepackage{subcaption}
\usepackage{indentfirst}
\usepackage[capitalize]{cleveref}
\crefname{section}{Sec.}{Secs.}
\Crefname{section}{Section}{Sections}
\Crefname{table}{Table}{Tables}
\crefname{table}{Tab.}{Tabs.}

\title{FractalAD: A simple industrial anomaly detection method using fractal anomaly generation and backbone knowledge distillation
}

\author{
  Xuan Xia$^1$, Weijie Lv$^2$, Xing He$^1$, Nan Li$^1$, Chuanqi Liu$^3$, Ning Ding$^1$\\
  1.Shenzhen Institute of Artificial Intelligence and Robotics for Society, Shenzhen\\
  2.Nanjing University of Aeronautics and Astronautics, Nanjing\\
  3.Shanghai Jiaotong University, Shanghai \\
  \texttt{\{xiaxuan, hexing, linan, dingning\}cuhk.edu.cn}\\
  \texttt{lvweijie@nuaa.edu.cn, chuanqil@sjtu.edu.cn}
}

\begin{document}
\maketitle
\setlength{\parindent}{2em}

\begin{abstract}
    Although industrial anomaly detection (AD) technology has made significant progress in recent years, generating realistic anomalies and learning priors of normal remain challenging tasks. In this study, we propose an end-to-end industrial anomaly detection method called FractalAD. Training samples are obtained by synthesizing fractal images and patches from normal samples. This fractal anomaly generation method is designed to sample the full morphology of anomalies. Moreover, we designed a backbone knowledge distillation structure to extract prior knowledge contained in normal samples. The differences between a teacher and a student model are converted into anomaly attention using a cosine similarity attention module. The proposed method enables an end-to-end semantic segmentation network to be used for anomaly detection without adding any trainable parameters to the backbone and segmentation head, and has obvious advantages over other methods in training and inference speed.. The results of ablation studies confirmed the effectiveness of fractal anomaly generation and backbone knowledge distillation. The results of performance experiments showed that FractalAD achieved competitive results on the MVTec AD dataset and MVTec 3D-AD dataset compared with other state-of-the-art anomaly detection methods.
\end{abstract}


\section{Introduction}
\label{sec:intro}

High-precision anomaly detection (AD) is of great significance in industrial defect detection from image data and provides an approach to product quality inspection that can be implemented rapidly without defect samples. In recent years, generative adversarial networks (GANs) \cite{venkataramanan2020attention, song2021anomaly}, flow models \cite{gudovskiy2022cflow, rudolph2022fully}, contrastive learning \cite{yi2020patch, zou2022spot}, knowledge distillation \cite{bergmann2020uninformed, wang2021student, deng2022anomaly} and other methods have significantly improved the performance of industrial image anomaly detection technologies. However, designing a general anomaly detection model with high accuracy and recall in various industrial image categories remains challenging.

Determining reasonable anomaly boundaries for detection models without abnormal samples has been actively researched \cite{xia2022gan}. In terms of available samples, anomaly detection models must solve the following two problems.

\textbf{Problem I}: Learn semantics, representation, and feature distribution of anomalies comprehensively without abnormal samples.

\textbf{Problem II}: Mine, generalize, and organize the prior knowledge of normal properly using normal samples.

\begin{figure*}[ht]
	\centering
		\includegraphics[width=0.8\textwidth]{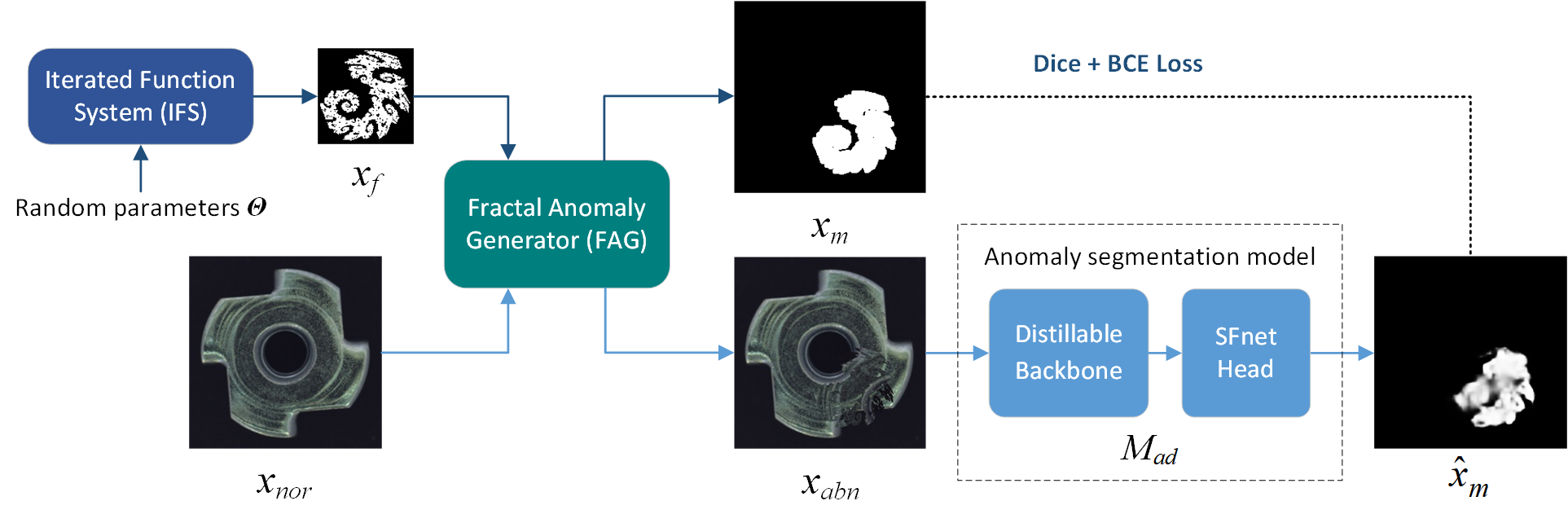}
	\caption{The training framework of FractalAD. Fractal anomaly generator (FAG) provides fractal anomaly image $x_{abn}$ which is beneficial to anomaly semantic learning for anomaly segmentation model $M_{ad}$. The distillable backbone provides the model $M_{ad}$ with prior knowledge contained in normal samples.}
	\label{fig:1}
\end{figure*}

Essentially, the solutions to these two problems are mutually exclusive, i.e., solving Problem I obviates the need to solve Problem II and vice versa. However, in practice, either problem is quite difficult to solve alone and thus must be considered together. For Problem I, several researchers have investigated self-supervised learning using simulated anomalies \cite{song2021anomaly, zou2022spot, zavrtanik2021draem, li2021cutpaste}, usually obtained by randomly sampling augmented patches from normal samples. Although simulating anomalies has proven to be an effective learning method, the types of simulation most suitable for industrial anomaly detection have not been definitively established in the literature.

Researchers have explored a variety of approaches for Problem II \cite{gudovskiy2022cflow, rudolph2022fully, defard2021padim, tsai2022multi}. Among these methods, knowledge distillation has gradually emerged because of its simplicity and efficiency \cite{bergmann2020uninformed, wang2021student, deng2022anomaly}. With the development of large learning models, knowledge distillation technology is worth further exploration.

In this study, to address these two problems, we propose an end-to-end industrial anomaly segmentation method called FractalAD. We aimed to achieve the most efficient anomaly detection using the simplest model and training method possible. As shown in \cref{fig:1}, the training framework design of FractalAD demonstrates our proposed solution to Problems I and II, which is to fully sample abnormal morphology through fractal anomaly images and extract prior knowledge contained in normal samples through the distillable backbone. Inspired by formula-driven supervised learning \cite{kataoka2020pre}, we use a series of data augmentation strategies to generate fractal anomaly images based on an iterated function system (IFS) \cite{anderson2022improving}. Fractal anomaly images exhibit far more diversity than other simulated anomalies owing to their potentially infinite morphological diversity, which is conducive to learning the semantics of anomalies caused by morphological changes. On the other hand, knowledge distillation of the backbone can extract prior knowledge contained in normal samples. Based on this approach, we adopt a simple strategy to generate anomaly attention to guide the model to detect more anomalies caused by non-morphological changes.

The contributions of this study are summarized as follows.

\begin{enumerate}
    \item{We propose a fractal anomaly generator (FAG) that uses fractal images to achieve full abnormal morphology sampling and improve the anomaly detection ability of a AD model.}
    
    \item{We design an optional distillable backbone and propose a cosine similarity attention module (CSAM). CSAM can convert the differences between backbones into a normalized spacial anomaly attention to improve the overall detection rate.}
    
    \item{We propose FractalAD as an anomaly detection framework with end-to-end training and inference. Based only on a simple semantic segmentation model, it does not include any modules with trainable parameters, and has obvious advantages over other methods in training and inference speed.}
    
    \item{The results of experimental evaluations showed that the performance of FractalAD was competitive with that of existing state-of-the-art methods on the MVTec AD dataset and MVTec 3D-AD dataset, thus demonstrating the effectiveness of FractalAD.}
\end{enumerate}

\section{Related work}
\label{sec:relwor}

\begin{figure*}[ht]
	\centering
		\includegraphics[width=0.8\textwidth]{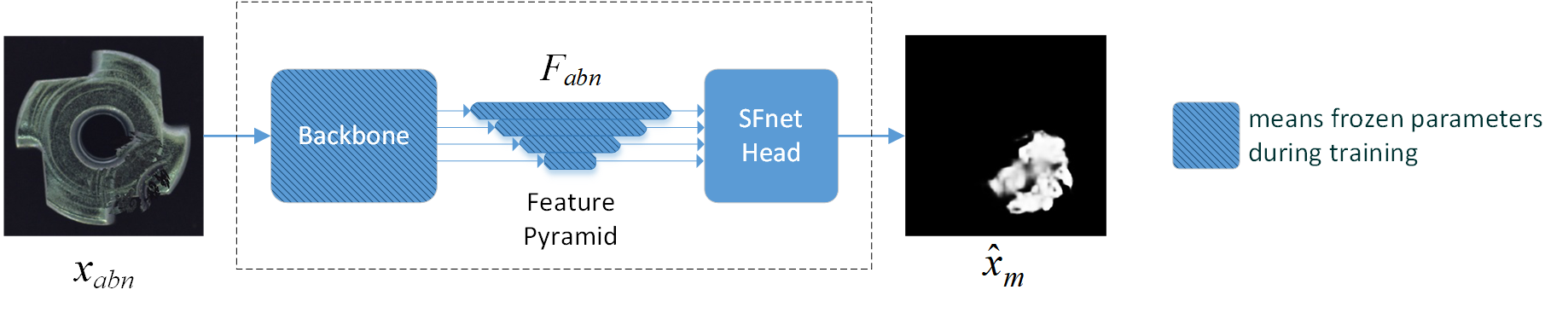}
	\caption{Basic structure of FractalAD model, which is a simple architecture of a backbone and an SFnet head \cite{li2020semantic}.}
	\label{fig:2}
\end{figure*}

\subsection{Industrial anomaly detection}
\label{section:2.1}

Industrial anomaly detection methods include image-level anomaly detection and pixel-level anomaly segmentation tasks. The demand for AD technology has increased with the development of the manufacturing industry, with many methods emerging in recent years.

As discussed in the introduction, self-supervised learning methods using simulated anomalies have been developed to learn anomalies without abnormal samples. For example, CutPaste\cite{li2021cutpaste} used random cut-and-paste-generated samples to learn a feature encoder beneficial for anomaly detection. DRAEM \cite{zavrtanik2021draem} uses Perlin noise to generate simulated abnormal samples to train an anomaly detection model. SPD \cite{zou2022spot} improved a data enhancement method in contrast learning and used the SimCLR \cite{chen2020simple} architecture to learn industrial image features and then detect and segment anomalies. However, the simulated anomalies generated by the existing methods differ significantly from real anomalies. Researchers usually only use the resultant output to train a stronger encoder; it cannot be directly used to train a model to segment anomalies.

For applications that assume abnormal samples are not available, using prior knowledge contained in normal samples has been explored. For example, CS-Flow \cite{rudolph2022fully} and AnoSeg \cite{song2021anomaly} used generative models to learn a distribution of normal features and discriminated anomalies in a latent space, whereas PaDiM \cite{defard2021padim} and PatchCore \cite{roth2022towards} extracted a feature pyramid of normal samples through pretrained models and discriminated anomalies according to the distribution boundary of the feature pyramid. Patch-SVDD \cite{yi2020patch} and InTra \cite{pirnay2022inpainting} distinguished anomalies by self-supervised learning of relationships between patches. Among these methods, the feature pyramid extraction method, represented by PatchCore\cite{roth2022towards}, has the simplest training framework and excellent detection performance. However, methods of this type require considerable memory resources, which affects their inference speed. In contrast, the recently developed knowledge distillation-based method \cite{bergmann2020uninformed, wang2021student, deng2022anomaly} also has a simple framework and relatively limited resource consumption. Thus, we consider this approach as showing significant potential.

\subsection{Formula-driven supervised learning}
\label{section:2.2}

The formula-driven supervised learning (FDSL) approach proposed in recent years is a specialized supervised learning method \cite{kataoka2021formula}. In contrast to natural image learning from the ImageNet dataset \cite{deng2009imagenet} or other natural images, FDSL does not rely on manual image acquisition and annotation. Instead, the model learns from images automatically generated using fractal geometry, computer graphics, and other methods. Existing studies \cite{kataoka2020pre, anderson2022improving, kataoka2021formula, nakashima2022can, kataoka2022replacing} have shown that such models can effectively learn representations through fractal images, Bessel curves \cite{farin2014curves}, and Perlin noise \cite{perlin2002improving}, improving the interpretability of the features and performing almost as well as pretrained models based on ImageNet.

At present, FDSL is mainly used to pretrain classification models and has rarely been implemented in AD. A rare example is DRAEM \cite{zavrtanik2021draem}, which uses Perlin noise to generate simulated anomalies. Perlin noise is a randomly generated texture noise. However, the anomalies it generated by the help of a third-party dataset and still differed considerably from real anomalies. In contrast, fractal images may be considered a better candidate. Owing to the butterfly effect, small perturbations of variables can cause unpredictable and infinite morphological changes in fractal images. Hence, the application of fractal images for anomaly detection is worth exploring.

\subsection{Knowledge-distillation-based AD}
\label{section:2.3}

Recently, knowledge distillation has been proposed for industrial anomaly detection using a student-teacher (S-T) framework. The student model is distilled using a pretrained teacher model on normal samples during the training phase. The discrepancy between the features generated by the student and the teacher is viewed as an anomaly when the input is abnormal. US \cite{bergmann2020uninformed} trains an ensemble of student networks on normal data at different scales, with both student and teacher architectures designed manually and identically. MKD \cite{salehi2021multiresolution}, which consists of a source network pretrained on ImageNet and a smaller cloner network, uses multi-scale feature alignment. In contrast, the teacher and student networks have the same architecture in STPM \cite{wang2021student}. The simple design of this framework makes its training and inference processes highly efficient.  Nevertheless, the same data flow in the S-T framework limits the capacity of the model. Thus, reverse distillation \cite{deng2022anomaly} was proposed to address this issue, in which the S-T model consists of a teacher encoder and a student decoder.

Although anomaly detection methods based on knowledge distillation are a topic of active research, to the best of our knowledge, no method to achieve end-to-end anomaly segmentation has been reported in the relevant literature. Therefore, considerable room for improvement remains.

\section{Method}
\label{sec:method}
\subsection{FractalAD}
\label{section:3.1}

\begin{figure*}[ht]
	\centering
		\includegraphics[width=0.8\textwidth]{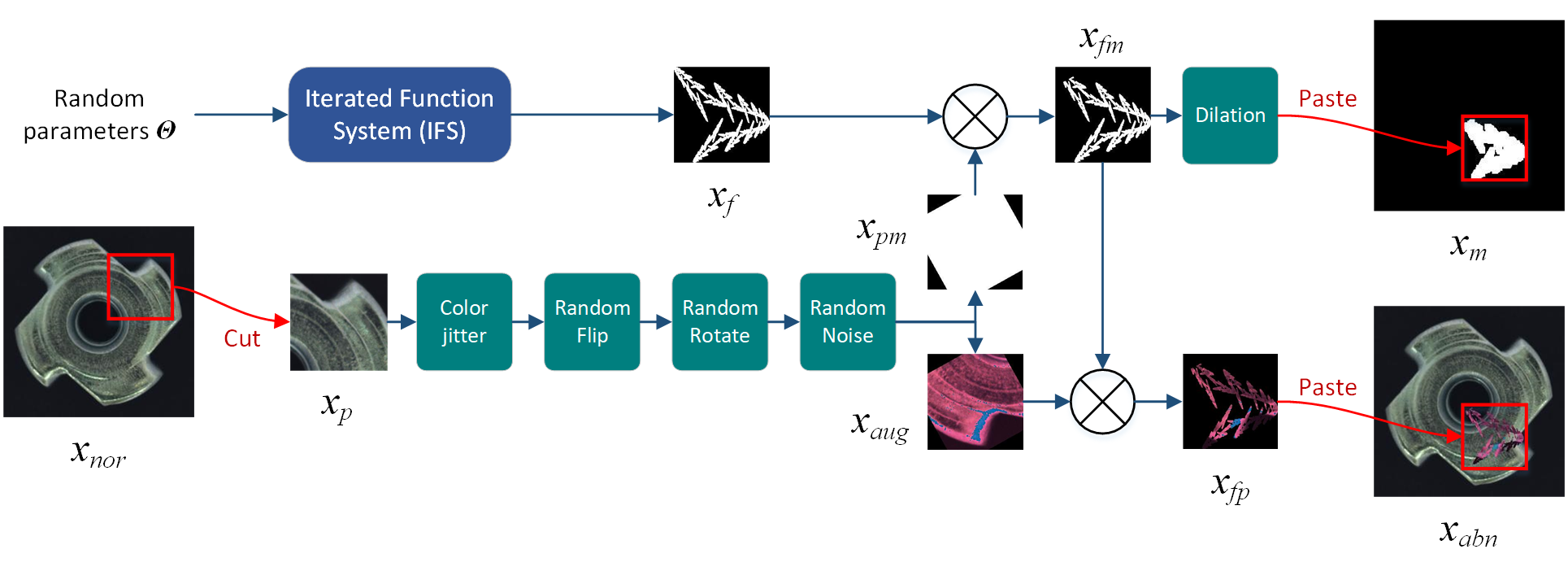}
	\caption{Process flow of fractal anomaly generation (FAG).}
	\label{fig:3}
\end{figure*}

The training framework for FractalAD is shown in \cref{fig:1}. One core idea of this framework is to generate abnormal samples by using random fractals to fully sample the abnormal morphology and facilitate the use of an end-to-end semantic segmentation model to locate anomalies. 
Mathematically, given training images $\boldsymbol{x}_{nor}$, we aim to generate synthetic abnormal images $\boldsymbol{x}_{abn}$ and their corresponding label masks $\boldsymbol{x}_m$ using fractal images $\boldsymbol{x}_f$ and a fractal anomaly generator (FAG)
\begin{equation}
  \boldsymbol{x}_{abn}, \boldsymbol{x}_{m}=\operatorname{F A G}\left(\boldsymbol{x}_{nor}, \boldsymbol{x}_{f}\right)
  \label{eq:1}
\end{equation}

Then, the anomaly detection model $M_{ad}$ is trained on these sample pairs, similar to a semantic segmentation model, by
\begin{equation}
  \hat{\boldsymbol{x}}_{m}=M_{ad}\left(\boldsymbol{x}_{abn}\right)
  \label{eq:2}
\end{equation}

We used the \textit{Dice+BCE} semantic segmentation loss function, which is a combination of two common losses
\begin{equation}
  L_{DB}=L_{Dice}+L_{BCE}
  \label{eq:3}
\end{equation}
where $L_{Dice}$ is \textit{Dice} loss \cite{milletari2016v}, and $L_{BCE}$ is the binary cross-entropy loss.

\cref{fig:2} shows the basic structure of the FractalAD model. For simplicity, we directly adopt an existing model composed of a backbone and an SFnet head \cite{li2020semantic}. Common models such as ResNet \cite{he2016deep} and EfficientNet \cite{tan2019efficientnet} can be adopted for the backbone. The SFnet head is responsible for fusing the semantic information in the feature pyramid output by the backbone and outputting the segmentation result of the anomaly, as given below.
\begin{equation}
\begin{aligned}
\hat{\boldsymbol{x}}_{m} &=\operatorname{SF}\left(\boldsymbol{F}_{a b n}\right) \\
&=\operatorname{SF}\left(B\left(\boldsymbol{x}_{a b n}\right)\right)
\end{aligned}
  \label{eq:4}
\end{equation}
where $B$ is the backbone, $\operatorname{SF}$ is SFnet head, and $\boldsymbol{F}_{abn}$ is the feature pyramid output of $B(\boldsymbol{x}_{a b n})$.

We simply chose the average value of $\hat{x}_{m}$ to measure the image-level anomaly score,

\begin{equation}
\begin{aligned}
s=\frac{1}{N M} \sum \hat{x}_{m}
\end{aligned}
  \label{eq:5}
\end{equation}
where $N \times M$ is the size of $\hat{x}_{m}$.

The flow alignment module (FAM) \cite{li2020semantic} in the SFnet head can efficiently align the feature semantics at different resolutions, which helps to detect anomalies at different scales. However, more importantly, it has relatively few parameters (only 0.33M for ResNet-18), which is conducive to a higher computational speed.

The advantage of FractalAD over other anomaly detection models is its simplicity. Our experimental results show that the simplicity of the model does not limit its anomaly detection performance. In the next section, we describe the design of a fractal anomaly generator used to train the proposed model.

\subsection{Fractal anomaly generation}
\label{section:3.2}

Fractal images can be generated through various mechanisms, such as an iterated function system (IFS), strange attractors, L-systems, and escape-time systems \cite{falconer2004fractal}. We used an IFS provided in \cite{anderson2022improving} to generate the fractal images $x_f$. We followed the default settings for this highly automated and efficient fractal image generator with a set of random parameters $\boldsymbol{\varTheta}$.
\begin{equation}
\begin{aligned}
\boldsymbol{x}_{f}=\operatorname{IFS}(\boldsymbol{\varTheta})
\end{aligned}
  \label{eq:6}
\end{equation}

On this basis, we constructed the process flow of the FAG, as shown in \cref{fig:3}. The process flow of the FAG is an improvement of the cut-paste strategy \cite{li2021cutpaste}. First, we cut out a patch $x_p$ from $x_{nor}$. Then, $x_{aug}$ and its corresponding mask $x_{pm}$ are obtained using multiple data augmentation strategies for $x_p$ (color jitter, random flipping, random rotation and random  noise). Through \textit{logical and} operation, a fractal image $x_f$ generated by IFS and $x_{pm}$ is synthesized into a fractal anomaly patch mask $x_{fm}$. Again, $x_{fm}$ and $x_{aug}$ are synthesized into fractal anomaly patch $x_{fp}$ by \textit{logical and} operation.  Finally, $x_{fp}$ is pasted onto $x_{nor}$ to generate the fractal anomaly image $x_{abn}$, and $x_{fm}$ is pasted to the same position on a black image to generate the fractal anomaly mask $x_m$ after dilation.

In the experiment, we found that the parameter settings of data augmentation and dilation directly affected the model's performance. For example, the angle range of random rotation must be limited, and dilation can significantly improve segmentation accuracy. The details are discussed in Section \ref{section:4.2} and Section \ref{section:4.4}. 

\subsection{Backbone knowledge distillation}
\label{section:3.3}

\begin{figure*}[t]
	\centering
		\includegraphics[width=0.8\textwidth]{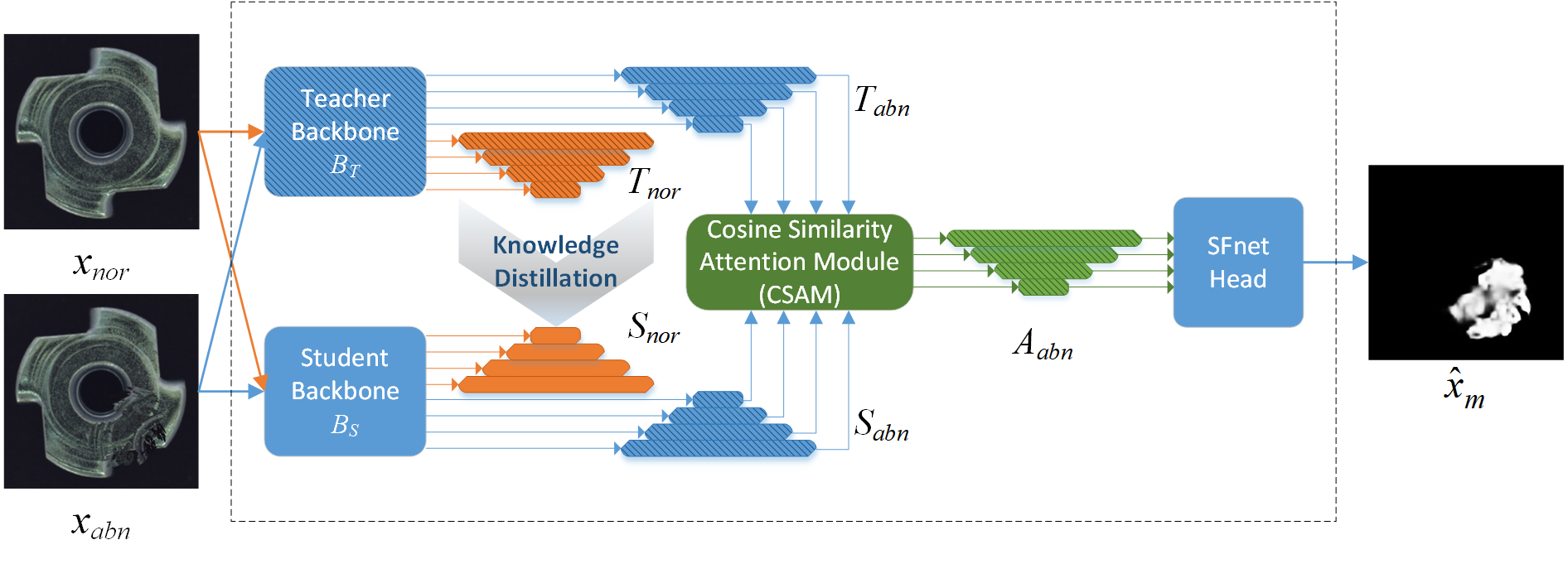}
	\caption{Upgraded model structure of FractalAD. In this structure, the student backbone $B_S$ is distilled by the teacher backbone $B_T$ (pre-trained on imagenet), and the cosine similarity attention module (CSAM) synthesizes the outputs of $B_T$ and $B_S$ into a new feature pyramid $\boldsymbol{A}_{abn}$. The orange path for knowledge distillation only exists in the training phase.}
	\label{fig:4}
\end{figure*}

Although fractal anomalies can realize sufficient sampling of defect morphology, they cannot realize sufficient sampling of different types of anomalies. The basic model of FractalAD, as shown in \cref{fig:2}, identifies an anomaly based on the difference in patches. However, some anomalies (such as cable swaps, shown in Figure 5.a) cannot form patch differences; therefore, the basic model cannot predict this type of anomaly without prior knowledge.

One solution is to use a memory bank \cite{defard2021padim, roth2022towards}. By directly storing the statistical characteristics of normal samples at different locations and scales, the model can perceive the location of abnormal occurrences by comparing differences in features. However, this approach typically requires considerable amount of memory resources and slows the model down significantly. Instead, we adopt a simpler knowledge distillation solution to obtain the difference between normal and abnormal conditions. Our solution was inspired by STPM \cite{wang2021student}, and we simplified it as an anomaly attention generation method. 

\cref{fig:4} shows the upgraded model structure of the FractalAD. It includes two backbones, the teacher backbone $B_T$ (pre-trained on imagenet) and the student backbone $B_S$, with the same structure. The feature pyramid $\boldsymbol{A}_{abn}$ input to SFnet is synthesized by the outputs $\boldsymbol{T}_{abn}$ and $\boldsymbol{S}_{abn}$ of both backbones through a \textit{cosine similarity attention module} (CSAM). Correspondingly, there are two gradient propagation paths. The blue path represents the anomaly prediction path, while the orange path represents the knowledge distillation path, which only exists in the training phase. We refer to the technique used here as backbone knowledge distillation (BKD).

The design of the structure is based on the following assumptions. Considering the student backbone can learn the domain prior knowledge contained in normal samples from the teacher backbone through knowledge distillation, we should be able to use the anomaly attention generated by them to guide SFnet to detect anomalies. 

Training the network remains simple. First, the knowledge distillation loss $L_{KD}$ is constructed between the feature pyramids $\boldsymbol{T}_{nor}=B_T(\boldsymbol{x}_{nor})$ and $\boldsymbol{S}_{nor}=B_S(\boldsymbol{x}_{nor})$ based on normal samples $\boldsymbol{x}_{nor}$
\begin{equation}
\begin{aligned}
L_{K D}=\sum_{i=1}^{4} E\left[1-\operatorname{CosSim}_{c h}\left(\boldsymbol{T}_{nor}^{i}, \boldsymbol{S}_{nor}^{i}\right)\right]
\end{aligned}
  \label{eq:7}
\end{equation}
where $\boldsymbol{T}_{nor}^{i}$ and $\boldsymbol{S}_{nor}^{i}$ are the $i^{th}$ feature maps of the $i^{th}$ stage of $B_T$ and $B_S$, respectively, and $\operatorname{CosSim}_{ch}$ is the function that computes the cosine similarity on the channel
\begin{equation}
\begin{aligned}
\operatorname{CosSim}_{c h}\left(\boldsymbol{T}_{nor}^{i}, \boldsymbol{S}_{nor}^{i}\right)=\frac{\left\langle \boldsymbol{T}_{nor}^{i}, \boldsymbol{S}_{nor}^{i}\right\rangle_{c h}}{\left\|\boldsymbol{T}_{nor}^{i}\right\|_{c h} \cdot\left\|\boldsymbol{S}_{nor}^{i}\right\|_{c h}}
\end{aligned}
  \label{eq:8}
\end{equation}

where $\langle \cdot , \cdot \rangle_{c h}$ computes the dot product on the channel and $\| \cdot \|_{c h}$ computes the second norm on the channel.

The semantic segmentation loss remains the same as that obtained from \cref{eq:3}. However, the feature pyramid $\boldsymbol{A}_{abn}$ in the SFnet head is generated through CASM
\begin{equation}
\begin{aligned}
\hat{\boldsymbol{x}}_{m} &=\operatorname{SF}\left(\boldsymbol{A}_{a b n}\right) \\
&=\operatorname{SF}\left(\operatorname{CSAM}\left(\boldsymbol{T}_{a b n}, \boldsymbol{S}_{a b n}\right)\right) \\
&=\operatorname{SF}\left(\operatorname{CSAM}\left(B_{T}\left(\boldsymbol{x}_{a b n}\right), B_{S}\left(\boldsymbol{x}_{a b n}\right)\right)\right)
\end{aligned}
  \label{eq:9}
\end{equation}
where
\begin{equation}
\begin{aligned}
\boldsymbol{A}_{a b n}^{i} &=\operatorname{CSAM}\left(\boldsymbol{T}_{a b n}^{i}, \boldsymbol{S}_{a b n}^{i}\right) \\
&=\operatorname{sg}\left(\boldsymbol{Att} \cdot \boldsymbol{S}_{a b n}^{i}\right) \\
&=\operatorname{sg}\left[\frac{1-\operatorname{CosSim}{ }_{c h}\left(\boldsymbol{T}_{a b n}^{i}, \boldsymbol{S}_{a b n}^{i}\right)}{2} \cdot \boldsymbol{S}_{a b n}^{i}\right]
\end{aligned}
  \label{eq:10}
\end{equation}
where $\operatorname{sg}(\cdot)$ denotes the stop-gradient operator, implying that the training of the feature pyramid $\boldsymbol{A}_{abn}$ is independent of the semantic segmentation task. $B_S$ only learns the knowledge of $\boldsymbol{x}_{nor}$ and does not participate in the training of anomaly segmentation. The CASM is a simple module that contains no learnable parameters.

From \cref{eq:9}, it may be observed that the CSAM serves to normalize the similarity of $\boldsymbol{T}_{a b n}^{i}$ and $\boldsymbol{S}_{a b n}^{i}$ to a normalized spacial anomaly attention $\boldsymbol{Att}$ and multiply with $\boldsymbol{S}_{a b n}^{i}$. $\boldsymbol{Att}$ indicates the degree of difference between  $\boldsymbol{T}_{a b n}$ and $\boldsymbol{S}_{a b n}$ to suppress the feature representation of normal regions, thus enabling the SFnet head to obtain a more accurate segmentation of abnormal regions.

To summarize, the total loss of the upgraded FractalAD is given as
\begin{equation}
\begin{aligned}
L_{total}=L_{D B}+L_{K D}
\end{aligned}
  \label{eq:11}
\end{equation}

Our experiments show that the basic FractalAD model performed well in some categories, and $L_{KD}$ is not required for training in all categories. Therefore, BKD is an alternative strategy in FractalAD. This point is discussed in more detail below in the description of the ablation studies.

\begin{table*}[t]
\centering
\scriptsize
\newcommand{\tabincell}[2]{\begin{tabular}{@{}#1@{}}#2\end{tabular}}
\begin{tabular}{|l|l|l|c|ll|}
\hline
\multicolumn{1}{|c|}{Module}  & \multicolumn{1}{c|}{Item}                   & \multicolumn{1}{c|}{Parameter}           & Value         & \multicolumn{2}{c|}{Descriptions}                                                                                                                               \\ \hline
\multirow{3}{*}{\tabincell{l}{Color \\ Jitter}} & \tabincell{l}{Brightness \& contrast \& \\ saturation \& hue} & Strength                                 & 0.2           & \multicolumn{1}{l|}{ColorJitter module of pytorch}                                                        & \multirow{3}{*}{\tabincell{l}{Randomly select \\ one of the 3 items}} \\ \cline{2-5}
                              & Brightness \& contrast                      & Strength                                 & 1             & \multicolumn{1}{l|}{ColorJitter module of pytorch}                                                        &                                                     \\ \cline{2-5}
                              & Random color                                & Strength range                           & {[}0,2{]}     & \multicolumn{1}{l|}{\tabincell{l}{Randomly select two channels and multiply \\ them by two random numbers between 0 and 2}} &                                                     \\ \hline
\multirow{2}{*}{\tabincell{l}{Random \\ Flip}}  & Right and left                              & Probability                              & 0.5           & \multicolumn{2}{l|}{\multirow{2}{*}{\tabincell{l}{Together with the random rotation module, this module realizes the \\ Angle transformation in four quadrants}}}                 \\ \cline{2-4}
                              & Top and bottom                              & Probability                              & 0.5           & \multicolumn{2}{l|}{}                                                                                                                                           \\ \hline
\tabincell{l}{Random \\ Rotate}                 & \multicolumn{1}{c|}{-}                      & Rotation angle range                     & {[}15°,75°{]} & \multicolumn{2}{l|}{\tabincell{l}{The angle close to 0° or 90° may cause the patch texture to be too \\ consistent with the surrounding}}                                         \\ \hline
\tabincell{l}{Random \\ Noise}                  & Gaussian noise                              & \tabincell{l}{Mean, standard deviation \\ and probability} & 0, 20, 0.5    & \multicolumn{2}{l|}{Add noise to the patch with a probability of 0.5}                                                                                           \\ \hline
Dilation                      & \multicolumn{1}{c|}{-}                      & Kernel size and iterations               & 3×3, 3        & \multicolumn{2}{l|}{\tabincell{l}{Dilation of mask is conducive to improving the detection rate of \\ small size anomalies}}                                                      \\ \hline
\end{tabular}
\caption{Parameter setting and corresponding description of fractal anomaly generator.}
\label{tab:1}
\end{table*}

\section{Experiments and analysis}
\label{sec:EA}

\subsection{Datasets, metric and Baselines}
\label{section:4.1}

We performed experiments on the MVTec AD dataset \cite{bergmann2019mvtec} and MVTec 3D-AD dataset \cite{bergmann2021mvtec}.  In the MVTec 3D-AD dataset, we only used RGB images. We evaluated the performance of our method at both the image and pixel levels by calculating the area under the receiver operating characteristics curve (AUROC). AUROC is a value between 0 and 1, with values close to 1 indicating better detection performance. We simply chose the average value of $\hat{x}_m$ to measure the image-level anomaly score for the testing images.

We compared our method with several state-of-the-art AD methods, including DRAEM \cite{zavrtanik2021draem}, CutPaste \cite{li2021cutpaste}, STPM \cite{wang2021student}, UniAD \cite{you2022unified}, RD \cite{deng2022anomaly}, PaDim \cite{defard2021padim}, PatchCore \cite{roth2022towards}, CFlow \cite{gudovskiy2022cflow}, CSflow \cite{rudolph2022fully}, FastFlow \cite{yu2021fastflow}, DifferNet \cite{rudolph2021same}, AST \cite{rudolph2023asymmetric}, EdgRec \cite{liu2022reconstruction}, and MSPBRL \cite{tsai2022multi}. For most of these methods, We used the results reported in the original studies for MVTec AD dataset. We used the results reported in AST \cite{rudolph2023asymmetric} and EdgRec \cite{liu2022reconstruction} for MVTec 3D-AD dataset.

\subsection{Training details and tricks}
\label{section:4.2}

FractalAD was trained using the AdamW \cite{loshchilov2017decoupled} optimizer with a batch size of 32. The size of the input image was set to 256 × 256 pixels. FractalAD was trained at an initial learning rate of 0.001 for 100 epochs for MVTec AD dataset and 200 epochs for MVTec 3D-AD dataset. The training set of each category expanded to 391 images for MVTec AD dataset (the number of images corresponding to the largest category in the training set in the MVTec AD dataset) by random replication to achieve the same number of training iterations for all categories for a fair comparison. FractalAD was implemented using the PyTorch 1.12\footnote{The code of FractalAD is available at https://github.com/AIRS-CSR/FractalAD.}. BKD was used only for object categories in MVTec AD dataset and for all categories in MVTec 3D-AD dataset. No changes were made to the backbone and SFnet head; therefore, the specific network structure is not described in detail. Training the FractalAD model took approximately 6$\sim$10 minutes per category of MVTec AD dataset on a single NVIDIA RTX 3090 GPU with a ResNet-18 backbone.

The setting of the FAG directly affects the anomaly generation quality. \cref{tab:1} lists the parameter settings of the fractal anomaly generator with descriptions. More detailed supplementary notes are described below.

\begin{enumerate}
    \item{Before the cut operation, threshold segmentation was performed on some object categories (\textit{capsule}, \textit{pill}, \textit{screw}, \textit{toothbrush} and \textit{zipper})  in MVTec AD dataset and  all categories in MVTec 3D-AD dataset to limit the cut-and-paste operations to the object area, ensuring that the fractal anomaly patch is located on the object.}

    \item{Using FAG is sufficient for texture categories (\textit{carpet, grid, leather, tile}, and \textit{wood}) in the MVTec AD dataset. BKD is not necessarily suitable for texture categories and may cause performance degradation (see ablation studies). We speculate that this may be because textures are repetitive image blocks and knowledge distillation cannot obtain meaningful high-level semantic information.}
    
    \item{Proper dilation of $x_m$ can guide the model to better segment anomalies, especially to improve the detection rate of subtle anomalies. The best parameter settings were determined through ablation studies.}
    
    \item{We conducted experiments to adjust the weights of the two losses in \cref{eq:11}, and found no significant effect on the results. So we are not going to show these ablation studies and the weights ware set to 1 by default.}
\end{enumerate}

\begin{table*}[ht]
\centering
\scriptsize
\setlength{\tabcolsep}{1.5mm}
\begin{tabular}{|cc|c|c|c|c|c|c|c|c|c|}
\hline
\multicolumn{2}{|c|}{category}                              & DREAM                & STPM        & UniAD                & RD                   & MSPBRL               & PatchCore   & CFlow                & FastFlow             & FractalAD            \\ \hline
\multicolumn{1}{|c|}{\multirow{5}{*}{texture}} & carpet     & 95.50/97.00          & 98.80/98.90 & 98.00/99.90          & 98.90/98.90          & 98.40/93.40          & 99.00/98.70 & 99.25/98.73          & \textbf{99.40/100.0} & 99.20/99.00          \\ \cline{2-11} 
\multicolumn{1}{|c|}{}                         & grid       & \textbf{99.70}/99.90 & 99.00/\textbf{100.0} & 94.60/98.50          & 99.30/\textbf{100.0}          & 98.50/\textbf{100.0}          & 98.70/98.20 & 98.99/99.60          & 98.30/99.70          & 98.50/98.75          \\ \cline{2-11} 
\multicolumn{1}{|c|}{}                         & leather    & 98.60/\textbf{100.0} & 99.30/99.90 & 98.30/\textbf{100.0}          & 99.40/\textbf{100.0}          & 99.10/99.30          & 99.30/\textbf{100.0} & \textbf{99.66/\textbf{100.0}} & 99.50/\textbf{100.0}          & 99.60/99.12          \\ \cline{2-11} 
\multicolumn{1}{|c|}{}                         & tile       & \textbf{99.20}/99.60 & 97.40/95.50 & 91.80/99.00          & 95.60/99.30          & 94.40/96.20          & 95.60/98.70 & 98.01/99.88          & 96.30/\textbf{100.0}          & 96.75/99.82          \\ \cline{2-11} 
\multicolumn{1}{|c|}{}                         & wood       & 96.40/99.10          & 97.20/99.20 & 93.40/97.90          & 95.30/99.20          & \textbf{97.50}/99.70 & 95.00/99.20 & 96.65/99.12          & 97.00/\textbf{100.0}          & 94.76/99.56          \\ \hline
\multicolumn{1}{|c|}{\multirow{10}{*}{object}} & bottle     & 99.10/99.20          & 98.80/\textbf{100.0} & 98.10/\textbf{100.0}          & 98.70/\textbf{100.0}          & 98.60/\textbf{100.0}          & 98.60/\textbf{100.0} & 98.98/\textbf{100.0}          & 97.70/\textbf{100.0}          & \textbf{99.15/\textbf{100.0}} \\ \cline{2-11} 
\multicolumn{1}{|c|}{}                         & cable      & 94.70/91.80          & 95.50/92.30 & 96.80/97.60          & 97.40/95.00          & 98.20/98.80          & \textbf{98.40}/99.50 & 97.64/97.59          & \textbf{98.40/\textbf{100.0}} & 97.30/98.78          \\ \cline{2-11} 
\multicolumn{1}{|c|}{}                         & capsule    & 94.30/98.50          & 98.30/88.00 & 97.90/85.30          & 98.70/96.30          & 97.90/97.20          & 98.80/98.10 & 98.98/97.68          & \textbf{99.10/\textbf{100.0}} & 98.24/96.77          \\ \cline{2-11} 
\multicolumn{1}{|c|}{}                         & hazelnut   & \textbf{99.70/\textbf{100.0}} & 98.50/\textbf{100.0} & 98.80/99.90          & 98.90/99.90          & 97.80/99.60          & 98.70/\textbf{100.0} & 98.89/99.98          & 99.10/\textbf{100.0}          & 98.07/\textbf{100.0}          \\ \cline{2-11} 
\multicolumn{1}{|c|}{}                         & metal nut  & \textbf{99.50}/98.70 & 97.60/\textbf{100.0} & 95.70/99.00          & 97.30/\textbf{100.0}          & 99.10/97.80          & 98.40/\textbf{100.0} & 98.56/99.26          & 98.50/\textbf{100.0}          & 98.79/97.46          \\ \cline{2-11} 
\multicolumn{1}{|c|}{}                         & pill       & 97.60/98.90          & 97.80/93.80 & 95.10/88.30          & 98.20/96.60          & 98.80/97.70          & 97.40/96.60 & 98.95/96.82          & \textbf{99.20/99.40} & 98.64/97.82          \\ \cline{2-11} 
\multicolumn{1}{|c|}{}                         & screw      & 97.60/93.90          & 98.30/88.20 & 97.40/91.90          & \textbf{99.60}/97.00 & 98.50/94.10          & 99.40/\textbf{98.10} & 98.86/91.89          & 99.40/97.80          & 98.65/95.14          \\ \cline{2-11} 
\multicolumn{1}{|c|}{}                         & toothbrush & 98.10/\textbf{100.0} & 98.90/87.80 & 97.80/95.00          & 99.10/99.50          & 99.00/\textbf{100.0}          & 98.70/\textbf{100.0} & 98.93/99.65          & 98.90/94.40          & \textbf{99.28/100.0} \\ \cline{2-11} 
\multicolumn{1}{|c|}{}                         & transistor & 90.90/93.10          & 82.50/93.70 & \textbf{98.70/100.0} & 92.50/96.70          & 97.70/98.90          & 96.30/\textbf{100.0} & 97.99/95.21          & 97.30/99.80          & 88.80/95.00          \\ \cline{2-11} 
\multicolumn{1}{|c|}{}                         & zipper     & 98.80/\textbf{100.0} & 98.50/93.60 & 96.00/96.70          & 98.20/98.50          & 98.60/99.50          & 98.80/99.40 & \textbf{99.08}/98.48          & 98.70/99.50          & 98.46/\textbf{100.0}          \\ \hline
\multicolumn{2}{|c|}{avg}                                   & 97.31/97.98          & 97.09/95.5  & 96.56/96.6           & 97.81/98.46          & 98.14/98.15          & 98.07/99.10 & \textbf{98.62}/98.26 & 98.45/\textbf{99.37}          & 97.62/98.48          \\ \hline
\end{tabular}
\caption{Comparison results on the MVTec AD dataset in terms of ROC-AUC \% with the format of pixel-level/image-level.}
\label{tab:2}
\end{table*}

\begin{table*}[ht]
\centering
\scriptsize
\begin{tabular}{|c|c|c|c|c|c|c|c|c|c|}
\hline
category     & PaDim & PatchCore & DifferNet & CSflow & AST           & FastFlow & CFlow         & EdgRec        & FractalAD      \\ \hline
bagel        & 97.5  & 87.6      & 85.9      & 94.1   & 94.7          & 89.3     & 88.0          & 89.7          & \textbf{97.62} \\ \hline
cable\_gland & 77.5  & 88.0      & 70.3      & 93.0   & 92.8          & 62.0     & 85.8          & 85.5          & \textbf{93.92} \\ \hline
carrot       & 69.5  & 79.1      & 64.3      & 82.7   & 85.1          & 79.5     & 82.8          & 88.5          & \textbf{91.64} \\ \hline
cookie       & 58.2  & 68.2      & 43.5      & 79.5   & \textbf{82.5} & 42.6     & 56.3          & 70.1          & 77.6           \\ \hline
dowel        & 95.9  & 91.2      & 79.7      & 99.0   & 98.1          & 88.0     & \textbf{98.6} & 94.5          & 98.26          \\ \hline
foam         & 66.3  & 70.1      & 79.0      & 88.6   & \textbf{95.1} & 72.8     & 73.8          & 82.1          & 85.87          \\ \hline
peach        & 85.8  & 69.5      & 78.7      & 73.1   & \textbf{89.5} & 65.1     & 75.7          & 67.9          & 84.25          \\ \hline
potato       & 53.5  & 61.8      & 64.3      & 47.1   & 61.3          & 56.0     & 62.8          & \textbf{75.5} & \textbf{70.26} \\ \hline
rope         & 83.2  & 84.1      & 71.5      & 98.6   & \textbf{99.2} & 98.2     & 97.0          & 96.8          & 96.97          \\ \hline
tire         & 76.0  & 70.2      & 59.0      & 74.5   & 82.1          & 61.3     & 72.0          & 81.8          & \textbf{93.66} \\ \hline
average      & 76.3  & 77.0      & 69.6      & 83.0   & 88.0          & 71.5     & 79.3          & 83.2          & \textbf{88.05} \\ \hline
\end{tabular}
\caption{Comparison results on the MVTec 3D-AD dataset in terms of image-level ROC-AUC \% (only RGB images).}
\label{tab:3}
\end{table*}

\begin{table}[]
\centering
\scriptsize
\setlength{\tabcolsep}{2mm}
\begin{tabular}{|cc|c|c|c|}
\hline
\multicolumn{2}{|c|}{module}            & \multirow{2}{*}{texture category} & \multirow{2}{*}{object category} & \multirow{2}{*}{average} \\ \cline{1-2}
\multicolumn{1}{|c|}{FAG}     & BKD     &                                   &                                  &                          \\ \hline
\multicolumn{1}{|c|}{without} & without & 90.71/97.65                       & 89.08/90.22                      & 89.62/92.70              \\ \hline
\multicolumn{1}{|c|}{with}    & without & \textbf{97.76/99.25}              & 90.87/93.98                      & 93.17/95.73              \\ \hline
\multicolumn{1}{|c|}{without} & with    & 91.68/90.61                       & 95.24/96.85                      & 94.06/94.77              \\ \hline
\multicolumn{1}{|c|}{with}    & with    & 97.67/93.92                       & \textbf{97.54/98.10}             & 97.60/96.01              \\ \hline
\multicolumn{2}{|c|}{default}           & \textbf{97.76/99.25}              & \textbf{97.54/98.10}             & \textbf{97.62/98.48}     \\ \hline
\end{tabular}
\caption{Evaluating the modules of FractalAD on the MVTec AD dataset in terms of ROC-AUC \% with the format of pixel-level/image-level (FAG: Fractal Anomaly Generation. BKD: Backbone Knowledge Distillation).}
\label{tab:4}
\end{table}

\subsection{Results and analysis}
\label{section:4.3}

\begin{figure}[ht]
  \centering
    \includegraphics[width=0.8\textwidth]{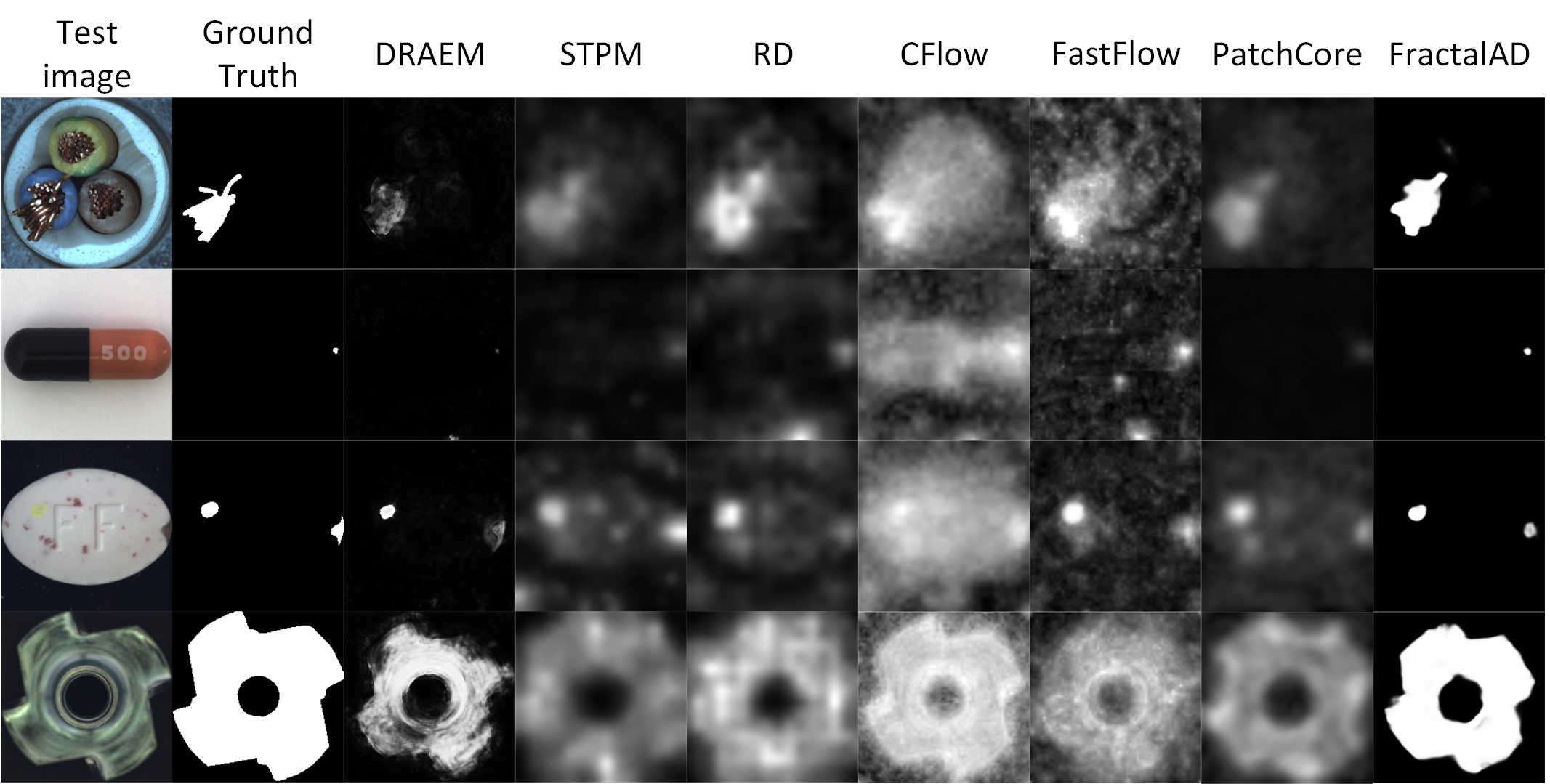}
    \caption{Visualization results of comparison of FractalAD and other methods.}
  \label{fig:5}
\end{figure}

\cref{tab:2} shows the comparison results between FractalAD and different methods on the MVTec AD dataset in terms of ROC-AUC \% at image and pixel levels. In the experiments, a ResNet-18 backbone was used in the FractalAD model. The comparison results showed that FractalAD achieved competitive results compared with other state-of-the-art AD methods. FractalAD achieved 98.48\% average image-level ROC-AUC. This results lagged behind PatchCore and FastFlow and superior to other methods.  FractalAD achieved an intermediate performance on average pixel-level ROC-AUC (97.46\%).

\cref{tab:3} shows the comparison results between FractalAD and the different methods on the MVTec 3D-AD dataset in terms of ROC-AUC \% at image level. The comparison results showed that FractalAD achieved best results (88.05\%) compared with other state-of-the-art AD methods. By observing the images in MVTec AD dataset and MVTec 3D-AD dataset, we can see that the objects in the MVTec 3D-AD dataset have more unknown shapes. This showed that fractal anomaly can help the model better adapt to shape changes.

\cref{fig:5} shows some visualization results (all methods show model output without threshold segmentation). FractalAD could segment abnormal regions more accurately and directly predict the mask of an abnormal region without any postprocessing. In contrast, the prediction results of DREAM easily produced artifacts, the prediction results of RD, STPM, CFlow, FastFlow and PatchCore were too vague. Although FractalAD is inferior to PatchCore and FastFlow in performance metrics, its results have better identifiability. More importantly, FractalAD does not need to use test sets and complex post-processing to determine the segmentation threshold, so it can be deployed quickly.

\subsection{Ablation studies}
\label{section:4.4}

\cref{tab:4} shows the results of the evaluation of the FractalAD modules on the MVTec AD dataset. The “without” of the FAG module means that fractal images are not used for generating abnormal patches (the cut-paste and data augmentation processes remain unchanged). The “without” of the BKD module means that the basic structure shown in \cref{fig:2} is used to train the model. Four combinations were evaluated in this study.  "default" refers to using only FAG for the texture categories, and FAG and BKD for the object categories.

From the table, the following conclusions can be drawn.

\begin{enumerate}
    \item{The FAG and BKD modules significantly improved the anomaly detection and segmentation abilities of the model, both being more effective together than individually.}
    
    \item{Relying only on the FAG module still significantly improved the performance of the model, demonstrating that fractal anomaly images can simulate anomalies better than conventional methods.}

    \item{The BKD module helped detect anomalies in object categories, demonstrating that it can support the model in identifying a normal prior. However, it may cause performance degradation. Therefore, we did not use the BKD module in the models for the texture categories in FractalAD as default.}
\end{enumerate}

\cref{tab:5} shows the experimental results of ablation studies of FAG module on Cutpaste and DRAEM. We used FAG to replace the anomaly generator in the original methods. The results showed that the performance of both methods has been improved after using the FAG module. This confirmed its generality and effectiveness.

\begin{table}[]
\centering
\scriptsize
\setlength{\tabcolsep}{2mm}
\begin{tabular}{|c|cccc|}
\hline
item           & \multicolumn{1}{c|}{img\_auc} & \multicolumn{1}{c|}{pixel\_auc} & \multicolumn{1}{c|}{img\_ap} & pixel\_ap      \\ \hline
cutpaste       & 89.2                          & -                               & -                            & -              \\
cutpaste(scar) & 90.6                          & -                               & -                            & -              \\
cutpaste+FAG   & \textbf{91.47}                & -                               & -                            & -              \\ \hline
DRAEM          & 97.98                         & 97.31                           & 0.99                         & \textbf{0.689} \\
DRAEM+FAG      & \textbf{98.51}                & \textbf{97.60}                  & 0.99                         & 0.669          \\ \hline
\end{tabular}
\caption{Experimental results of ablation studies of FAG module on Cutpaste and DRAEM (img\_auc: ROC-AUC \% of image-level, pixel\_auc: ROC-AUC \% of pixel-level, img\_ap: average precision of image-level, pixel\_ap: average precision of pixel-level).}
\label{tab:5}
\end{table}

\begin{table}[t]
\scriptsize
\centering
\begin{tabular}{|c|c|}
\hline
Backbone                                              & metric                                \\ \hline
Resnet18                                              & \textbf{97.62/98.48}                  \\ \hline
Resnet34                                              & 97.51/97.60                           \\ \hline
Resnet50                                              & 96.37/97.16                           \\ \hline
WideResnet50 \cite{zagoruyko2016wide}                 & 97.07/97.34                           \\ \hline
Convnext (tiny) \cite{liu2022convnet}                 & 95.13/96.90                           \\ \hline
Efficientnet v2 (tiny) \cite{tan2021efficientnetv2}   & 93.21/96.09                           \\ \hline
swin (tiny) \cite{liu2021swin}                        & 95.54/97.58                           \\ \hline
\end{tabular}
\caption{Experimental results with different backbones on the MVTec AD dataset (ROC-AUC \% of pixel-level/image-level).}
\label{tab:6}
\end{table}

\begin{table}[t]
\scriptsize
\centering
\setlength{\tabcolsep}{1mm}
\begin{tabular}{|c|c|c|c|c|c|}
\hline
iterations & 0           & 1           & 2           & 3                    & 4           \\ \hline
metric     & 93.33/97.65 & 95.95/98.40 & 96.09/98.50 & \textbf{97.62/98.48} & 97.41/98.39 \\ \hline
\end{tabular}
\caption{Experimental results of using different iterations of the fractal anomaly mask dilation on the MVTec AD dataset (ROC-AUC \% of pixel-level/image-level).}
\label{tab:7}
\end{table}

\begin{table}[t]
\scriptsize
\centering
\setlength{\tabcolsep}{1mm}
\begin{tabular}{|l|l|l|}
\hline
method                     & \multicolumn{1}{c|}{time(ms)} & \multicolumn{1}{c|}{FPS} \\ \hline
PatchCore                  & 28.23                         & 35.41                    \\ \hline
CFlow                      & 27.75                         & 36.04                    \\ \hline
DRAEM                      & 20.18                         & 49.56                    \\ \hline
FastFlow(Wide-ResNet50-2)* & 17.25                         & 57.95                    \\ \hline
FastFlow(ResNet18)*        & 13.53                         & 73.94                    \\ \hline
RD                         & 17.16                         & 58.28                    \\ \hline
STPM                       & 12.31                         & 81.21                    \\ \hline
FractalAD (w BKD)          & \textbf{10.63}                & \textbf{94.04}           \\ \hline
FractalAD (w/o BKD)        & \textbf{5.03}                 & \textbf{198.89}          \\ \hline
\end{tabular}
\caption{Experimental results of inference speed comparison ($256 \times 256$ size on RTX 3090).}
\label{tab:8}
\end{table}

\cref{tab:6} shows the impact of different backbones on the detection performance. The experimental results showed that increasing the complexity of the backbone did not improve model performance, possibly because  industrial images are simpler than natural images, and overly complex backbone is not necessary for the segmentation head. Using ResNet-18, the simplest option, proved to be the best approach.

\cref{tab:7} shows the experimental results obtained using different iterations of fractal anomaly mask dilation. The kernel size of dilation was 3×3. Multiple iterations are equivalent to using a larger kernel. The experimental results show that three iterations achieved the best performance; therefore, we adopted this value as the default setting for FractalAD.

\cref{tab:8} shows the experimental results of inference speed comparison. The results demonstrated the speed advantage of FractalAD. Because it uses a simple semantic segmentation network and does not require any post-processing, FractalAD can run at a faster speed than other methods.

\subsection{Conclusion}

In this study, we proposed a simple industrial anomaly detection method using fractal anomaly generation and backbone knowledge distillation. The experimental results have confirmed that fractal anomaly images are useful for simulating real anomalies, and the backbone knowledge distillation can provide prior knowledge for anomaly segmentation. More importantly, this research has shown that anomaly detection task can be achieved by simply training a segmentation model end-to-end. This means that the model can gain great advantages over other methods in training and inference speed. Therefore, although the performance of our approach still has room for improvement, we believe it has the potential to be widely applied with further development. In future research, we plan to develop more comprehensive anomaly generation methods and more accurate anomaly detection and segmentation technologies.

\bibliographystyle{unsrt}  
\bibliography{references}

\end{document}